\documentclass[10pt,twocolumn,letterpaper]{article}

\usepackage[T1]{fontenc}
\usepackage{cvpr}

\usepackage{multirow}
\usepackage{colortbl}
\usepackage{array}

\definecolor{cvprblue}{rgb}{0.21,0.49,0.74}
\usepackage[breaklinks,colorlinks,allcolors=cvprblue]{hyperref}

\def\confName{CVPR}
\def\confYear{2026}

\definecolor{zerorecall}{HTML}{F44336}
\newcommand{\pp}{\text{pp}}
\newcommand{\best}[1]{\textbf{#1}}
\newcommand{\sigmark}[1]{$^{#1}$}
\newcommand{\redzero}{\textcolor{zerorecall}{\textbf{0.0}}}

\title{Beyond Small-Loss: Rethinking Noise-Robust Training for Frozen Vision Foundation Models in Medical Image Classification}

\author{
Zitong Li$^*$ \qquad Haoyu Wang$^*$\\
Department of Biostatistics and Health Informatics\\
King's College London\\
{\tt\small zitong.2.li@kcl.ac.uk \qquad haoyu.7.wang@kcl.ac.uk}\\
$^*$Equal contribution
}

\begin{document}
\maketitle

\begin{abstract}
Vision Foundation Models (VFMs) with frozen feature extractors and lightweight classification heads have become the dominant paradigm for medical image classification. However, label noise---ubiquitous in clinical annotations---poses a significant challenge to this paradigm. We systematically investigate whether existing noise-robust training methods, which rely on the \emph{small-loss criterion} to distinguish clean from noisy samples, remain effective under frozen VFM settings. Our analysis reveals that this criterion fundamentally fails: loss distributions of clean and noisy samples overlap by 53--61\%, rendering loss-based sample selection unreliable. We adapt a prediction-agreement cascade that exploits the inherent stability of frozen feature spaces, where prediction consistency provides a more reliable signal than loss magnitude. Through comprehensive experiments on ISIC2019 (8-class, 54:1 imbalance) and BloodMNIST (8-class, balanced) with three VFM backbones (DINOv2, BiomedCLIP, ResNet50), we uncover two critical findings: (1)~Co-Teaching catastrophically collapses under class imbalance with asymmetric noise, dropping recall to 0\% for three minority classes while maintaining 68\% overall accuracy; (2)~method rankings statistically significantly reverse between symmetric and asymmetric noise ($p{<}0.005$), challenging the notion of a universal noise-robust method. Our results provide concrete guidelines for practitioners deploying frozen VFMs on noisy medical data.
\end{abstract}

\section{Introduction}
\label{sec:intro}

Vision Foundation Models (VFMs) such as DINOv2~\cite{oquab2024dinov2} and BiomedCLIP~\cite{zhang2024biomedclip} are transforming medical image analysis. The dominant deployment paradigm extracts features from a frozen VFM backbone and trains a lightweight classification head (\eg a two-layer MLP), avoiding the computational cost of full fine-tuning while achieving competitive performance~\cite{zhang2024foundation_survey}. This ``extract once, train many'' approach is especially attractive in clinical settings where computational resources are limited and reproducibility is paramount.

However, medical image annotations are inherently noisy. Inter-observer disagreement rates of 10--40\% are well-documented in dermatology~\cite{tschandl2018ham10000} and pathology, making noise-robust training essential. Over the past six years, a rich literature has developed methods for learning with noisy labels---Co-Teaching~\cite{han2018coteaching}, DivideMix~\cite{li2020dividemix}, ELR~\cite{liu2020elr}, and most recently CUFIT~\cite{yu2024cufit}---all built upon a common foundation: the \emph{small-loss criterion}~\cite{arpit2017memorization}, which assumes that samples with smaller training loss are more likely to have correct labels.

A critical but unexamined question arises: \textbf{does the small-loss criterion remain valid when training only a lightweight MLP on frozen VFM features?} This setting fundamentally differs from end-to-end training or adapter fine-tuning, where the backbone parameters co-evolve with the loss landscape. Under frozen features, the MLP has far fewer learnable parameters, producing compact loss distributions that may not separate clean from noisy samples.

In this work, we provide the first systematic investigation of this question. Our contributions are:

\begin{enumerate}
  \item \textbf{Failure of the small-loss criterion.} We demonstrate that under frozen VFM settings, loss distributions of clean and noisy samples overlap by 53--61\% (vs.\ 20--30\% in adapter fine-tuning), rendering loss-based sample selection unreliable. This finding holds across three noise conditions (symmetric 20\%, symmetric 40\%, asymmetric 40\%).

  \item \textbf{Prediction-agreement cascade.} We adapt the cascade architecture from CUFIT~\cite{yu2024cufit}, replacing loss ranking with prediction agreement as the sample selection signal. Ablation studies show the cascade progressively improves performance, with the first filtering stage contributing up to +10.8$\pp$ balanced accuracy.

  \item \textbf{Co-Teaching collapse under imbalance.} We reveal that Co-Teaching catastrophically fails on imbalanced medical data with asymmetric noise: three minority classes drop to 0\% recall while overall accuracy remains 68\%. This collapse is consistent across all three VFM backbones and absent on balanced data (BloodMNIST), confirming it is driven by class imbalance, not noise alone.

  \item \textbf{Method ranking reversal.} We show that the relative ranking of noise-robust methods \emph{statistically significantly} reverses between noise types: ELR outperforms prediction-agreement cascade under symmetric noise ($p{=}0.005$), while the cascade outperforms ELR under asymmetric noise ($p{=}0.001$). This has direct practical implications for method selection.
\end{enumerate}

\section{Related Work}
\label{sec:related}

\paragraph{Learning with noisy labels.}
The small-loss criterion~\cite{arpit2017memorization,natarajan2013learning} underpins most noise-robust methods. Co-Teaching~\cite{han2018coteaching} trains two networks that teach each other using small-loss samples. DivideMix~\cite{li2020dividemix} fits a Gaussian Mixture Model to per-sample losses for clean/noisy separation, then applies semi-supervised learning. ELR~\cite{liu2020elr} takes a different approach, regularizing the model to maintain early-learning predictions rather than selecting samples. CUFIT~\cite{yu2024cufit} introduces a curriculum cascade for VFM fine-tuning with adapter modules. A critical gap remains: \emph{all these methods were validated under end-to-end training or adapter fine-tuning}; their effectiveness under frozen MLP settings is unknown.

\paragraph{Vision Foundation Models in medical imaging.}
DINOv2~\cite{oquab2024dinov2} learns robust visual features through self-supervised training, and frozen linear probing is its standard evaluation protocol. BiomedCLIP~\cite{zhang2024biomedclip} provides domain-specific representations for biomedical images. The ``frozen features + lightweight head'' paradigm is rapidly becoming the default deployment strategy~\cite{zhang2024foundation_survey}, yet almost no work has studied noise robustness in this specific setting.

\paragraph{Class imbalance and noisy labels.}
The interaction between class imbalance and label noise is theoretically understood to be problematic: Gui \etal~\cite{gui2021towards} showed that the small-loss criterion requires the noise transition matrix to be diagonally dominant, while Zhang \etal~\cite{zhang2023rcal} showed that assumptions behind noisy-label methods can break down under long-tailed distributions. Our work provides an \emph{empirical} validation of these concerns in the frozen VFM + medical imaging setting.

\section{Why Small-Loss Fails Under Frozen VFMs}
\label{sec:small_loss}

\subsection{Experimental Setup}
\label{sec:setup}

\paragraph{Architecture.} We use a consistent architecture across all experiments: a frozen VFM backbone extracts features, which are fed to a two-layer MLP ($d_\text{feat} \to 256 \to K$) with BatchNorm, ReLU, and dropout (0.3). Training uses Adam with cosine annealing (lr=$10^{-3}$), class-weighted cross-entropy loss, and early stopping (patience=7).

\paragraph{Backbones.} We evaluate three VFMs: DINOv2-Base (ViT-B/14, $d{=}768$), BiomedCLIP (ViT-B/16, $d{=}512$), and ImageNet-pretrained ResNet50 ($d{=}2048$).

\paragraph{Datasets.} (1)~\textbf{ISIC2019}~\cite{tschandl2018ham10000,codella2019isic}: 25,331 dermoscopic images across 8 diagnostic categories with extreme class imbalance (NV: 12,875 vs.\ DF: 239, ratio ${\approx}$54:1). (2)~\textbf{BloodMNIST}~\cite{yang2023medmnist}: 17,092 blood cell microscopy images across 8 cell types with moderate imbalance (ratio ${\approx}$5:1).

\paragraph{Noise injection.} We apply two noise types at rates $\eta \in \{0\%, 10\%, 20\%, 30\%, 40\%\}$: \emph{symmetric noise} flips labels uniformly at random; \emph{asymmetric noise} flips labels to semantically similar classes. For ISIC2019, confusion pairs follow dermatological similarity (\eg MEL$\to$NV, BCC$\to$BKL). For BloodMNIST, pairs follow cell morphology: granulocytes interchange (BAS$\leftrightarrow$NEU$\leftrightarrow$EOS), and mononuclear cells interchange (LYM$\leftrightarrow$MON), reflecting realistic annotation confusion in hematology.

\paragraph{Evaluation.} All results report balanced accuracy (BalAcc)---the mean of per-class recalls---to account for class imbalance. We use 5 independent runs with paired $t$-tests and Cohen's $d$ for statistical significance.

\subsection{Loss Distribution Analysis}
\label{sec:loss_dist}

The small-loss criterion assumes that clean samples produce systematically lower losses than noisy samples, enabling reliable separation. We test this assumption by training a standard MLP classifier on noisy data and comparing the per-sample loss distributions of known-clean and known-noisy samples (using the ground truth available in synthetic noise experiments).

\begin{table}[t]
\centering
\caption{Loss distribution statistics under frozen DINOv2 features. The \textbf{Overlap} column measures the overlap coefficient between clean and noisy loss distributions. Values above 50\% indicate that a majority of samples cannot be correctly classified by loss ranking alone.}
\label{tab:loss_overlap}
\small
\begin{tabular}{@{}lccc@{}}
\toprule
\textbf{Condition} & \textbf{Clean Loss} & \textbf{Noisy Loss} & \textbf{Overlap} \\
\midrule
Sym.\ 20\% & $1.36 \pm 0.66$ & $2.23 \pm 0.88$ & 53.2\% \\
Sym.\ 40\% & $1.58 \pm 0.56$ & $2.12 \pm 0.62$ & 60.7\% \\
Asym.\ 40\% & $1.23 \pm 0.63$ & $1.62 \pm 0.80$ & 61.0\% \\
\bottomrule
\end{tabular}
\end{table}

\begin{figure}[t]
\centering
\includegraphics[width=\linewidth]{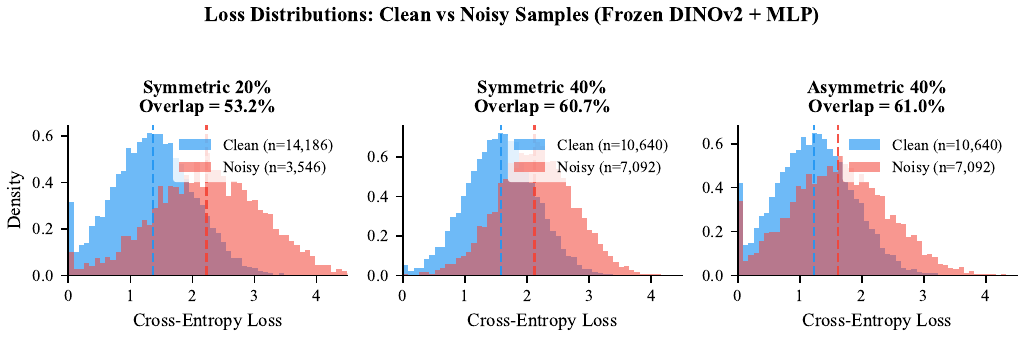}
\caption{Loss distributions of clean (blue) and noisy (red) samples under frozen DINOv2 features. The overlap coefficient increases from 53\% to 61\% as noise becomes more severe or asymmetric, indicating that loss ranking cannot reliably distinguish clean from noisy samples.}
\label{fig:loss_dist}
\end{figure}

\cref{tab:loss_overlap} and \cref{fig:loss_dist} reveal that the overlap between clean and noisy loss distributions ranges from \textbf{53.2\% to 61.0\%} under frozen DINOv2 features. The Kolmogorov-Smirnov test confirms the distributions differ ($p{<}0.001$), but the practical separability is poor: more than half of all samples fall in the overlapping region where loss alone cannot distinguish clean from noisy.

Two mechanisms drive this high overlap. First, the frozen backbone constrains the MLP to a low-dimensional parameter space, producing compact loss distributions with limited dynamic range. Second, asymmetric noise creates label flips between semantically similar classes (\eg MEL$\to$NV), whose feature representations already overlap in the frozen feature space, making their losses inherently similar regardless of label correctness.

Notably, the overlap \emph{increases} with noise severity and with asymmetric noise structure. At 40\% asymmetric noise---the most clinically realistic scenario---61\% of samples are inseparable by loss. This stands in stark contrast to adapter fine-tuning settings, where the trainable backbone creates more dispersed loss landscapes with typically 20--30\% overlap~\cite{yu2024cufit,li2020dividemix}.

\subsection{Frozen Feature Space Stability}
\label{sec:stability}

The failure of loss-based selection points to a need for alternative signals. We observe a key property of frozen VFM settings: \textbf{the feature space is identical regardless of label noise}. Since the backbone is frozen, features are extracted once and remain fixed---noise affects only the label-feature alignment, not the feature geometry itself. We quantify this: the average intra-class feature distance changes by only 2.4\% from clean to 40\% noise (26.32$\to$26.95 in $\ell_2$ norm), while inter-class distance decreases from 10.88 to 4.45 due to label corruption. The feature geometry is stable; it is the label-feature alignment that degrades.

This stability has a powerful implication: the prediction of a classifier trained on the features directly reflects the agreement between the learned decision boundary and the feature space structure. If a sample's predicted label matches its given label, this agreement suggests the given label is consistent with the feature space geometry.

\paragraph{Quantitative comparison of selection signals.}
To directly compare loss ranking and prediction agreement as clean sample detectors, we compute their precision and recall for identifying truly clean samples (\cref{tab:selection}).

\begin{table}[t]
\centering
\caption{Clean sample detection: loss ranking vs.\ prediction agreement. Precision = fraction of selected samples that are truly clean. At matched selection rates, loss achieves higher precision under symmetric noise, but its advantage \emph{shrinks by $4\times$} under asymmetric noise. Prediction agreement is more robust across noise types ($-$3$\pp$ vs.\ $-$13$\pp$ precision drop from sym to asym).}
\label{tab:selection}
\small
\begin{tabular}{@{}llccc@{}}
\toprule
\textbf{Condition} & \textbf{Method} & \textbf{Prec.} & \textbf{Rec.} & \textbf{Sel.\%} \\
\midrule
\multirow{2}{*}{Sym 40\%} & Pred.\ agree. & 82.2\% & 53.3\% & 38.9\% \\
                           & Loss (matched) & \best{95.2\%} & 61.7\% & 38.9\% \\
\midrule
\multirow{2}{*}{Asym 40\%} & Pred.\ agree. & 79.2\% & 53.7\% & 40.7\% \\
                            & Loss (matched) & 82.2\% & 55.8\% & 40.7\% \\
\midrule
\multicolumn{2}{@{}l}{\textit{Precision drop (sym$\to$asym)}} & & & \\
\quad & Pred.\ agree. & $-$3.0$\pp$ & & \\
\quad & Loss ranking & $-$13.0$\pp$ & & \\
\bottomrule
\end{tabular}
\end{table}

\cref{tab:selection} reveals a nuanced picture. In a single-shot comparison, loss ranking achieves higher precision under symmetric noise (95.2\% vs.\ 82.2\%). However, under asymmetric noise---the clinically more realistic scenario---the gap shrinks from 13$\pp$ to just 3$\pp$. Critically, prediction agreement is \textbf{$4\times$ more robust} to the noise type: its precision drops by only 3$\pp$ from symmetric to asymmetric noise, compared to a 13$\pp$ drop for loss ranking. At practical selection rates (${\sim}$60\%), loss precision degrades further to 68.4\% under asymmetric noise, meaning nearly 1-in-3 selected ``clean'' samples are actually noisy. This robustness advantage, combined with the threshold-free nature of prediction agreement (no need to estimate noise rate), makes it a more reliable signal for the cascade framework, particularly when the noise structure is unknown.

\section{Prediction-Agreement Cascade}
\label{sec:method}

\subsection{Method Design}

Motivated by the failure of loss-based selection and the stability of frozen features, we adapt the cascade architecture of CUFIT~\cite{yu2024cufit} with prediction agreement as the selection criterion. The cascade operates in three stages:

\paragraph{Stage 1: Linear Probing Module (LPM).}
A linear classifier $f_1: \mathbb{R}^d \to \mathbb{R}^K$ is trained on all samples using class-weighted cross-entropy. The simplicity of linear classification limits overfitting to noise. Samples where $\arg\max f_1(\mathbf{x}) = y$ (prediction matches given label) are retained as a ``likely clean'' subset $\mathcal{S}_1$.

\paragraph{Stage 2: Intermediate Adapter Module (IAM).}
A two-layer MLP $f_2$ is trained on $\mathcal{S}_1$. Since $\mathcal{S}_1$ is cleaner than the full dataset, $f_2$ learns a more accurate decision boundary. A second round of prediction-agreement filtering produces an even cleaner subset $\mathcal{S}_2 \subseteq \mathcal{S}_1$.

\paragraph{Stage 3: Last Adapter Module (LAM).}
The final MLP $f_3$ is trained on $\mathcal{S}_2$ and used for inference. Each stage trains on progressively cleaner data, enabling increasingly accurate classification.

\paragraph{Implementation details.}
All three modules are trained \emph{jointly} for 50 epochs with shared data batches. At each training step, agreement masks are computed dynamically: IAM receives gradients only from LPM-agreed samples, and LAM only from IAM-agreed samples. Under symmetric 40\% noise, LPM retains approximately 39\% of training samples (${\sim}$6,900 of 17,732), and IAM further refines this to approximately 35\%. The selection rate adapts automatically to the noise level without requiring noise rate estimation.

\subsection{Key Difference from CUFIT}

The original CUFIT~\cite{yu2024cufit} was designed for adapter fine-tuning, where the backbone is partially trainable and loss distributions are well-separated. It uses \emph{loss ranking} for sample selection: samples with the smallest losses are retained at each stage. In our frozen MLP setting, we replace loss ranking with \emph{prediction agreement}: samples whose predicted labels match their given labels are retained. This substitution is motivated by the high loss overlap (53--61\%) demonstrated in \cref{sec:loss_dist}, which makes loss ranking unreliable, and by the feature space stability (\cref{sec:stability}), which makes prediction agreement a trustworthy signal.

\subsection{Ablation Study}

\begin{table}[t]
\centering
\caption{Ablation of cascade stages on ISIC2019 with DINOv2. \textbf{Only LPM}: linear classifier only. \textbf{LPM$+$LAM}: LPM filtering followed by MLP training (skip IAM). \textbf{Full}: complete three-stage cascade. $\Delta$ denotes the gain from adding the first filtering stage (Only LPM $\to$ LPM$+$LAM).}
\label{tab:ablation}
\small
\setlength{\tabcolsep}{2pt}
\begin{tabular}{@{}lcccc@{}}
\toprule
\textbf{Config.} & \textbf{Clean} & \textbf{Sym 20\%} & \textbf{Sym 40\%} & \textbf{Asym 40\%} \\
\midrule
Only LPM & 62.2 & 47.2 & 40.9 & 48.1 \\
LPM$+$LAM & \best{66.6} & 56.6 & 51.7 & 56.5 \\
Full Cascade & 66.0 & \best{58.4} & \best{51.9} & \best{56.7} \\
\midrule
$\Delta$ LPM$\to$LAM & +4.4 & +9.4 & +10.8 & +8.4 \\
$\Delta$ LAM$\to$Full & $-$0.6 & +1.8 & +0.2 & +0.2 \\
\bottomrule
\end{tabular}
\end{table}

\begin{figure}[t]
\centering
\includegraphics[width=\linewidth]{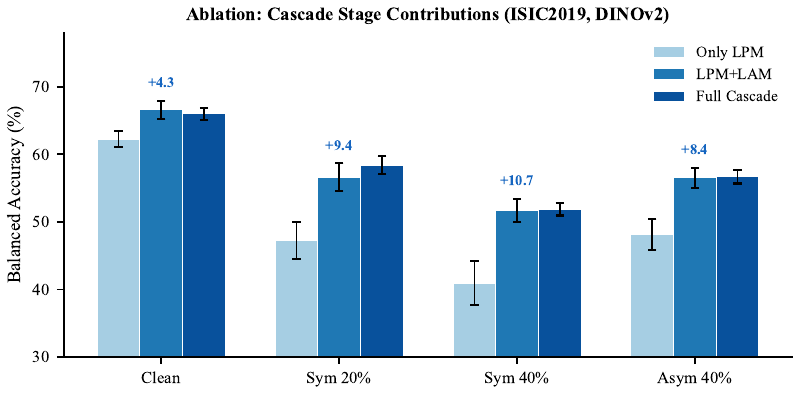}
\caption{Ablation of cascade stages. The first filtering stage (Only LPM $\to$ LPM+LAM) yields the largest gain (+4.4 to +10.8$\pp$), which grows with noise severity. Error bars: $\pm$1 std over 5 runs.}
\label{fig:ablation}
\end{figure}

\cref{tab:ablation} and \cref{fig:ablation} validate the cascade design. The first filtering stage (LPM$\to$LPM$+$LAM) contributes the majority of the improvement, with gains of \textbf{+4.4} to \textbf{+10.8}$\pp$ BalAcc. The benefit increases with noise severity, confirming that prediction-agreement filtering effectively removes noisy samples. The second filtering stage (IAM) provides smaller but consistent gains (+0.2 to +1.8$\pp$), suggesting that in frozen settings, a two-stage cascade captures most of the benefit. This is a practical finding: \textbf{practitioners can achieve most of the noise robustness with a simple two-stage pipeline}.

\section{Experiments}
\label{sec:experiments}

\subsection{Compared Methods}

We evaluate eight methods: \textbf{Baseline CE} (cross-entropy with class weights), \textbf{Linear Probe} (single linear layer), \textbf{Label Smoothing}~\cite{muller2019labelsmoothing}, \textbf{Co-Teaching}~\cite{han2018coteaching}, \textbf{DivideMix}~\cite{li2020dividemix}, \textbf{CUFIT} (our prediction-agreement cascade, adapted from~\cite{yu2024cufit}), \textbf{SOP} (self-paced oversampling), and \textbf{ELR}~\cite{liu2020elr}.

\subsection{Main Results on ISIC2019}

\begin{table*}[t]
\centering
\caption{Balanced accuracy (\%) on ISIC2019 with DINOv2 backbone. \textbf{Bold}: best per condition. Significance markers: \sigmark{**}$p{<}0.01$, \sigmark{*}$p{<}0.05$ vs.\ Baseline CE (paired $t$-test, 5 runs). The ranking of CUFIT and ELR \emph{reverses} between symmetric and asymmetric noise.}
\label{tab:main_isic}
\small
\begin{tabular}{@{}lcccccccc@{}}
\toprule
\textbf{Condition} & \textbf{CE} & \textbf{Lin.\ Probe} & \textbf{Lab.\ Smooth.} & \textbf{Co-Teach.} & \textbf{DivideMix} & \textbf{CUFIT} & \textbf{SOP} & \textbf{ELR} \\
\midrule
\multicolumn{9}{@{}l}{\textit{Symmetric Noise}} \\
\quad Clean     & 68.4\tiny{$\pm$0.8} & 62.2\tiny{$\pm$1.2} & \best{69.3}\tiny{$\pm$1.3} & 63.2\tiny{$\pm$0.8} & 62.1\tiny{$\pm$3.3} & 66.0\tiny{$\pm$0.9} & 67.7\tiny{$\pm$1.4} & 63.4\tiny{$\pm$1.6} \\
\quad $\eta{=}10\%$ & 58.3\tiny{$\pm$2.5} & 52.3\tiny{$\pm$1.8} & 59.9\tiny{$\pm$1.3} & \best{63.2}\tiny{$\pm$1.0} & 59.1\tiny{$\pm$1.7} & 61.5\tiny{$\pm$1.4} & 59.7\tiny{$\pm$1.5} & 59.5\tiny{$\pm$1.9} \\
\quad $\eta{=}20\%$ & 53.9\tiny{$\pm$1.2} & 47.2\tiny{$\pm$2.7} & 54.2\tiny{$\pm$0.7} & \best{59.2}\tiny{$\pm$1.5} & 56.5\tiny{$\pm$1.1} & 58.4\sigmark{*}\tiny{$\pm$1.3} & 53.2\tiny{$\pm$2.8} & 58.0\tiny{$\pm$1.4} \\
\quad $\eta{=}30\%$ & 48.8\tiny{$\pm$2.8} & 43.0\tiny{$\pm$4.9} & 49.0\tiny{$\pm$1.7} & 56.1\tiny{$\pm$0.9} & 53.6\tiny{$\pm$2.9} & 55.7\sigmark{**}\tiny{$\pm$1.4} & 51.4\tiny{$\pm$0.9} & \best{56.0}\tiny{$\pm$0.8} \\
\quad $\eta{=}40\%$ & 45.4\tiny{$\pm$0.8} & 40.9\tiny{$\pm$3.3} & 46.3\tiny{$\pm$1.2} & 50.5\tiny{$\pm$2.3} & 49.7\tiny{$\pm$2.2} & 51.9\sigmark{**}\tiny{$\pm$1.0} & 46.1\tiny{$\pm$1.5} & \best{55.6}\sigmark{**}\tiny{$\pm$1.0} \\
\midrule
\multicolumn{9}{@{}l}{\textit{Asymmetric Noise}} \\
\quad $\eta{=}10\%$ & 63.8\tiny{$\pm$1.0} & 58.8\tiny{$\pm$2.5} & \best{66.3}\tiny{$\pm$1.0} & 59.4\tiny{$\pm$2.7} & 61.0\tiny{$\pm$2.1} & 64.0\tiny{$\pm$1.9} & 64.3\tiny{$\pm$1.3} & 60.8\tiny{$\pm$0.9} \\
\quad $\eta{=}20\%$ & 61.2\tiny{$\pm$1.2} & 57.4\tiny{$\pm$1.7} & 62.0\tiny{$\pm$1.7} & 46.8\tiny{$\pm$3.3} & 58.2\tiny{$\pm$1.2} & \best{61.7}\tiny{$\pm$1.2} & 60.0\tiny{$\pm$2.1} & 58.7\tiny{$\pm$0.8} \\
\quad $\eta{=}30\%$ & 55.9\tiny{$\pm$2.1} & 51.5\tiny{$\pm$1.9} & 57.9\tiny{$\pm$1.4} & 39.1\tiny{$\pm$1.3} & 55.4\tiny{$\pm$2.6} & \best{57.9}\sigmark{*}\tiny{$\pm$1.1} & 55.8\tiny{$\pm$1.1} & 56.6\tiny{$\pm$0.7} \\
\quad $\eta{=}40\%$ & 53.6\tiny{$\pm$1.3} & 48.1\tiny{$\pm$2.3} & 55.1\tiny{$\pm$2.4} & 35.1\tiny{$\pm$1.8} & 52.1\tiny{$\pm$3.5} & \best{56.7}\sigmark{**}\tiny{$\pm$1.0} & 53.0\tiny{$\pm$1.0} & 53.3\tiny{$\pm$1.3} \\
\bottomrule
\end{tabular}
\end{table*}

\cref{tab:main_isic} presents the full results on ISIC2019 with DINOv2. Several key findings emerge:

\paragraph{Prediction-agreement cascade consistently outperforms CE under noise.}
CUFIT significantly outperforms the CE baseline at $\eta{\geq}20\%$ for both noise types (all $p{<}0.03$, paired $t$-test). The advantage grows with noise severity: +4.5$\pp$ at symmetric 20\% vs.\ +6.5$\pp$ at symmetric 40\%, and +3.1$\pp$ at asymmetric 40\% ($p{=}0.003$, $d{=}3.0$).

\begin{figure}[t]
\centering
\includegraphics[width=\linewidth]{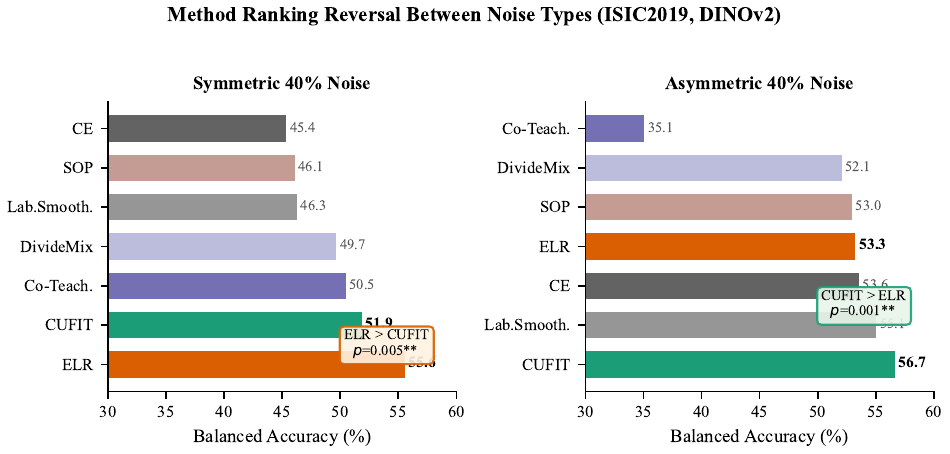}
\caption{Method ranking reversal between noise types. Under symmetric noise (left), ELR ranks first; under asymmetric noise (right), CUFIT ranks first. Both differences are statistically significant ($p{<}0.005$). Co-Teaching collapses to last place under asymmetric noise.}
\label{fig:reversal}
\end{figure}

\paragraph{Method ranking reverses between noise types (\cref{fig:reversal}).}
Under \textbf{symmetric 40\%} noise, ELR achieves the highest BalAcc (\best{55.6}\%) and significantly outperforms CUFIT (51.9\%, $p{=}0.005$, $d{=}{-}2.6$). However, under \textbf{asymmetric 40\%} noise, CUFIT achieves \best{56.7}\% and significantly outperforms ELR (53.3\%, $p{=}0.001$, $d{=}3.7$). This \emph{statistically significant ranking reversal} has direct implications: \textbf{there is no universal best noise-robust method}---the optimal choice depends on the noise structure.

\paragraph{Co-Teaching collapses under asymmetric noise + imbalance.}
Co-Teaching performs competitively under symmetric noise (50.5\% at $\eta{=}40\%$), but catastrophically fails under asymmetric noise, dropping to 35.1\% at $\eta{=}40\%$---significantly worse than even the CE baseline ($p{<}10^{-5}$, $d{=}13.3$). We analyze this collapse in detail in \cref{sec:collapse}.

\subsection{Co-Teaching Collapse Analysis}
\label{sec:collapse}

\begin{table}[t]
\centering
\caption{Per-class recall (\%) on ISIC2019 under asymmetric 40\% noise (DINOv2). Classes ordered by training set size. \redzero{} indicates complete failure on that class. Co-Teaching achieves high recall on majority classes (NV, BCC) but completely abandons three minority classes.}
\label{tab:perclass}
\small
\begin{tabular}{@{}lr|cccc@{}}
\toprule
\textbf{Class} & \textbf{$N$} & \textbf{CE} & \textbf{CUFIT} & \textbf{ELR} & \textbf{Co-T.} \\
\midrule
NV   & 6705 & 67.4 & 58.2 & 77.7 & \best{82.5} \\
BCC  & 3323 & 48.6 & 52.9 & 33.3 & \best{83.9} \\
MEL  & 1113 & 44.2 & 55.8 & 52.4 & \best{56.6} \\
BKL  & 1099 & 39.4 & 38.4 & 13.0 & \best{44.7} \\
AK   &  867 & \best{54.9} & 50.2 & 59.4 & 13.2 \\
SCC  &  628 & 52.6 & \best{55.5} & 51.9 & \redzero{} \\
DF   &  253 & 50.0 & \best{57.2} & 52.8 & \redzero{} \\
VASC &  253 & 71.6 & 85.3 & \best{85.8} & \redzero{} \\
\midrule
\textit{Range} && 32.2 & 46.9 & 72.8 & 83.9 \\
\textit{BalAcc} && 53.6 & \best{56.7} & 53.3 & 35.1 \\
\bottomrule
\end{tabular}
\end{table}

\begin{figure}[t]
\centering
\includegraphics[width=\linewidth]{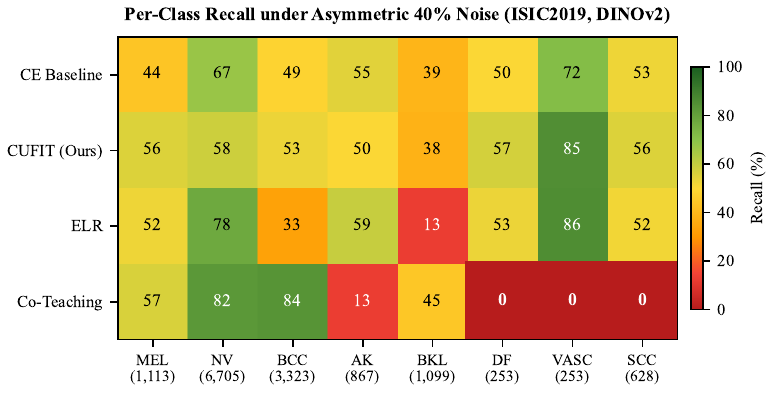}
\caption{Per-class recall heatmap under asymmetric 40\% noise (ISIC2019, DINOv2). Co-Teaching completely abandons DF, VASC, and SCC (recall = 0\%, dark red), while achieving high recall on majority classes. CUFIT provides the most balanced recall distribution.}
\label{fig:perclass}
\end{figure}

\cref{tab:perclass} and \cref{fig:perclass} expose the mechanism behind Co-Teaching's collapse. While Co-Teaching achieves \emph{high} recall on majority classes (NV: 82.5\%, BCC: 83.9\%), it completely abandons three minority classes (\textbf{DF, VASC, SCC: all 0\% recall}) and nearly abandons a fourth (AK: 13.2\%). Its overall accuracy of 68.1\% appears reasonable, but the balanced accuracy of 35.1\% reveals the true picture.

The collapse results from a triple interaction:
\begin{enumerate}
  \item \textbf{Extreme class imbalance} (${\sim}$54:1) means majority-class samples naturally dominate the low-loss region.
  \item \textbf{Asymmetric noise} disproportionately corrupts specific minority classes (\eg SCC$\to$AK, DF$\to$BKL), further inflating their losses.
  \item \textbf{Global small-loss selection} in Co-Teaching then systematically excludes these high-loss minority samples, effectively removing entire classes from training.
\end{enumerate}

\noindent\textbf{Cross-dataset validation of the collapse mechanism.}
On BloodMNIST (balanced, ${\sim}$5:1 imbalance), Co-Teaching achieves 94.1\% BalAcc under asymmetric 40\% noise---no collapse occurs. This confirms that the catastrophic failure is driven by \textbf{class imbalance interacting with asymmetric noise}, not by asymmetric noise alone.

\noindent\textbf{Clinical significance.} In dermatological screening, DF (dermatofibroma), VASC (vascular lesions), and SCC (squamous cell carcinoma) being completely missed is clinically dangerous---particularly SCC, which is malignant and requires treatment. A model reporting 68\% accuracy while being blind to three diagnostic categories would create a false sense of reliability.

\paragraph{Fairness across classes.}
\cref{tab:perclass} also reveals differences in class fairness (measured by the range between best and worst per-class recall). CUFIT achieves the most \emph{balanced} performance across classes (range: 46.9$\pp$), while Co-Teaching is the most unbalanced (83.9$\pp$). CE, despite lower overall BalAcc, is relatively balanced (32.2$\pp$ range).

\subsection{Cross-Dataset and Cross-Backbone Validation}

\begin{table}[t]
\centering
\caption{Cross-dataset validation on BloodMNIST (balanced, DINOv2). All methods perform well on clean data; differences emerge under noise. Co-Teaching does \textbf{not} collapse on balanced data.}
\label{tab:bloodmnist}
\small
\begin{tabular}{@{}lcccc@{}}
\toprule
\textbf{Condition} & \textbf{CE} & \textbf{CUFIT} & \textbf{ELR} & \textbf{Co-T.} \\
\midrule
Clean     & 98.5\tiny{$\pm$0.1} & 98.1\tiny{$\pm$0.3} & 98.1\tiny{$\pm$0.1} & 98.1\tiny{$\pm$0.1} \\
Sym.\ 20\%  & 92.9\tiny{$\pm$0.6} & 97.7\tiny{$\pm$0.3} & \best{98.0}\tiny{$\pm$0.1} & 97.1\tiny{$\pm$0.2} \\
Sym.\ 40\%  & 89.4\tiny{$\pm$1.9} & 96.9\tiny{$\pm$0.4} & \best{97.3}\tiny{$\pm$0.3} & 94.7\tiny{$\pm$1.4} \\
Asym.\ 40\% & 82.3\tiny{$\pm$4.2} & 94.6\tiny{$\pm$0.2} & \best{96.3}\tiny{$\pm$0.1} & 94.1\tiny{$\pm$0.6} \\
\bottomrule
\end{tabular}
\end{table}

\begin{table}[t]
\centering
\caption{Cross-backbone validation on ISIC2019 at asymmetric 40\% noise. Co-Teaching collapse is consistent across all three backbones.}
\label{tab:crossbackbone}
\small
\setlength{\tabcolsep}{4pt}
\begin{tabular}{@{}lccccc@{}}
\toprule
\textbf{Backbone} & \textbf{CE} & \textbf{CUFIT} & \textbf{ELR} & \textbf{Co-T.} & \textbf{$\Delta$\scriptsize{(CU$-$CE)}} \\
\midrule
DINOv2      & 53.6 & \best{56.7} & 53.3 & 35.1 & +3.1 \\
BiomedCLIP  & \best{51.8} & 51.5 & 50.5 & 32.3 & $-$0.3 \\
ResNet50    & 46.0 & \best{52.7} & 50.7 & 31.5 & +6.7 \\
\bottomrule
\end{tabular}
\end{table}

\begin{figure}[t]
\centering
\includegraphics[width=\linewidth]{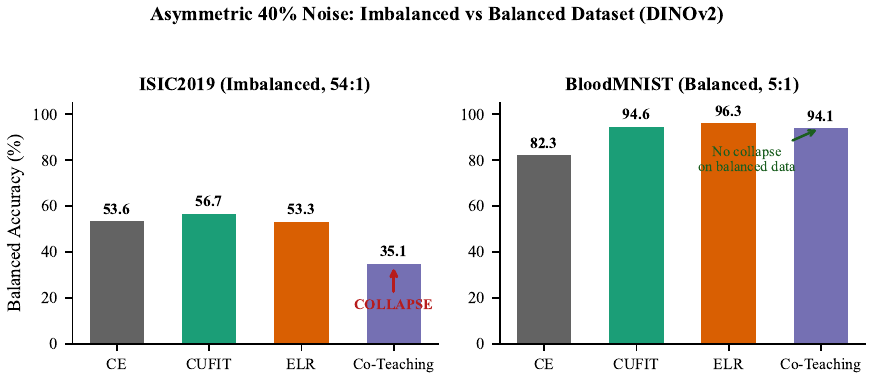}
\caption{Co-Teaching collapse is driven by class imbalance, not noise alone. Under identical asymmetric 40\% noise, Co-Teaching collapses on imbalanced ISIC2019 (left, BalAcc=35.1\%) but performs well on balanced BloodMNIST (right, BalAcc=94.1\%).}
\label{fig:crossdataset}
\end{figure}

\paragraph{BloodMNIST results (\cref{tab:bloodmnist} and \cref{fig:crossdataset}).}
On the balanced BloodMNIST dataset, all noise-robust methods substantially outperform the CE baseline under noise (up to +14.0$\pp$ for ELR at asymmetric 40\%). Crucially, Co-Teaching performs well (94.1\%), confirming that its ISIC2019 collapse is specific to the imbalanced + asymmetric noise interaction.

Two cross-dataset patterns are noteworthy:
\begin{itemize}
  \item \textbf{ELR is consistently the top performer on balanced data}, achieving the highest BalAcc across all noise conditions on BloodMNIST.
  \item \textbf{CUFIT's advantage over CE is consistent}: +6.5$\pp$ (ISIC2019 sym 40\%) and +7.5$\pp$ (BloodMNIST sym 40\%).
\end{itemize}

\paragraph{Cross-backbone consistency (\cref{tab:crossbackbone}).}
The Co-Teaching collapse under asymmetric noise is \emph{backbone-independent}: BalAcc drops to 35.1\% (DINOv2), 32.3\% (BiomedCLIP), and 31.5\% (ResNet50). CUFIT outperforms or matches CE on all backbones, with gains ranging from $-$0.3$\pp$ (BiomedCLIP) to +6.7$\pp$ (ResNet50). The largest gains appear on the weakest backbone, suggesting that prediction-agreement filtering is most beneficial when features are less discriminative.

\subsection{Method Selection Guide}

\begin{table}[t]
\centering
\caption{Practical method selection guide based on comprehensive experimental evidence across 2 datasets, 3 backbones, and 5 noise rates.}
\label{tab:guide}
\small
\begin{tabular}{@{}p{2.8cm}lp{3.0cm}@{}}
\toprule
\textbf{Data Characteristics} & \textbf{Rec.} & \textbf{Evidence} \\
\midrule
Balanced + any noise  & ELR & Best on BloodMNIST; stable across conditions \\
Imbal.\ + sym.\ noise & ELR & $>$ CUFIT on ISIC ($p{=}0.005$) \\
Imbal.\ + asym.\ noise & CUFIT & $>$ ELR ($p{=}0.001$); most balanced recall \\
Low noise / clean & Lab.\ Sm. & Simple; best on clean ISIC \\
\midrule
\multicolumn{3}{@{}l}{\textbf{Avoid:} Co-Teaching on imbalanced data (collapse risk)} \\
\bottomrule
\end{tabular}
\end{table}

\cref{tab:guide} synthesizes our findings into actionable recommendations. The central message is that \textbf{noise type matters}: the optimal method depends on whether the noise is symmetric or asymmetric. When the noise structure is unknown, CUFIT is a safer choice for imbalanced data due to its most balanced per-class recall distribution, while ELR is preferred for balanced data.

\section{Discussion}
\label{sec:discussion}

\paragraph{Frozen features vs.\ adapter fine-tuning.}
Our results show that the small-loss criterion, which is effective under adapter fine-tuning~\cite{yu2024cufit}, fails under frozen features. This suggests a fundamental trade-off: adapter fine-tuning enables loss-based noise detection but requires more computation, while frozen features are efficient but require alternative noise detection signals. The prediction-agreement approach bridges this gap, achieving noise robustness without backbone modification.

\paragraph{The danger of aggregate metrics.}
Co-Teaching's asymmetric noise results highlight a critical evaluation issue: overall accuracy (68.1\%) masks the complete failure on three classes. In medical AI, where every diagnostic category matters, \textbf{balanced accuracy and per-class recall should be mandatory metrics}. We urge the community to adopt these as standard practice for imbalanced medical datasets.

\paragraph{Fairness implications.}
Noise-robust methods are implicitly assumed to improve performance uniformly. Our per-class analysis shows this is not the case: methods that improve aggregate BalAcc may dramatically worsen performance on specific (typically minority) classes. Future work should explicitly optimize for worst-class recall or other fairness metrics.

\paragraph{Limitations.}
(1) Our noise is synthetic; real clinical annotation noise has unknown structure. However, the ranking reversal between noise types suggests our conclusions generalize to mixed-noise scenarios. (2) We evaluate only the frozen MLP setting; adapter and full fine-tuning may behave differently. (3) While ISIC2019 and BloodMNIST provide contrasting characteristics (imbalanced vs.\ balanced, dermoscopy vs.\ microscopy), additional datasets would strengthen generalizability. (4) Our approach assumes the noise rate is approximately known for Co-Teaching's forget rate parameter; in practice, this must be estimated.

\section{Conclusion}
\label{sec:conclusion}

We have demonstrated that the small-loss criterion---the foundation of most noise-robust training methods---fails under frozen Vision Foundation Model settings due to high loss distribution overlap (53--61\%). By replacing loss ranking with prediction agreement, we achieve significant improvements over the CE baseline across two medical imaging datasets and three VFM backbones. Our analysis reveals that Co-Teaching catastrophically collapses on imbalanced data with asymmetric noise (three classes at 0\% recall), and that no single noise-robust method dominates across all noise types ($p{<}0.005$). These findings provide concrete, evidence-based guidelines for practitioners deploying frozen VFMs on noisy medical imaging data. Code and pretrained feature extractors will be released to facilitate reproducibility.

{
    \small
    \bibliographystyle{ieeenat_fullname}
    \bibliography{references}
}


\end{document}